# Automated Generation of Connectionist Expert Systems For Problems Involving Noise and Redundancy

Stephen I. Gallant*

College of Computer Science
Northeastern University
Boston, Ma. 02115 USA

June 4, 1987

**Abstract**

When creating an expert system, the most difficult and expensive task is constructing a knowledge base. This is particularly true if the problem involves noisy data and redundant measurements.

This paper shows how to modify the MACIE process for generating connectionist expert systems from training examples so that it can accommodate noisy and redundant data. The basic idea is to dynamically generate appropriate training examples by constructing both a 'deep' model and a noise model for the underlying problem. The use of winner-take-all groups of variables is also discussed.

These techniques are illustrated with a small example that would be very difficult for standard expert system approaches.

*TOPICS/KEYWORDS:* Machine Learning, Connectionist Models, Knowledge acquisition, Expert Systems, MACIE

## 1 Introduction

There is widespread agreement that the most difficult, time consuming, and expensive task in constructing an expert system is extracting rules from a human expert and debugging the resulting knowledge base. If noise or redundant data is present the problem is even more difficult; for some situations there is no expert.

In response to this problem there has been extensive work on extracting rules from sets of training examples. See [Michalski 1983, 1986], the journal *Machine Learning*, and proceedings of the annual workshops in machine learning for details. (See also [Gallant 1986a].)

Another type of approach involves adapting probabilistic or fuzzy methods to expert systems. See [Cheeseman 1983, 1984; Pearl 1985, 1986; Zadeh 1965] and the *International Journal of Intelligent Systems* for some of this work.

A third approach involves remembering training examples and emulating the closest one when presented with a new instance. See [Stanfill & Waltz 1986] for a recent hardware intensive implementation of this approach.

*Partially supported by National Science Foundation grant IRI-8611596 and a grant from the Northeastern University Research and Scholarship Development Fund. Thanks to Mark Frydenberg for helpful comments.

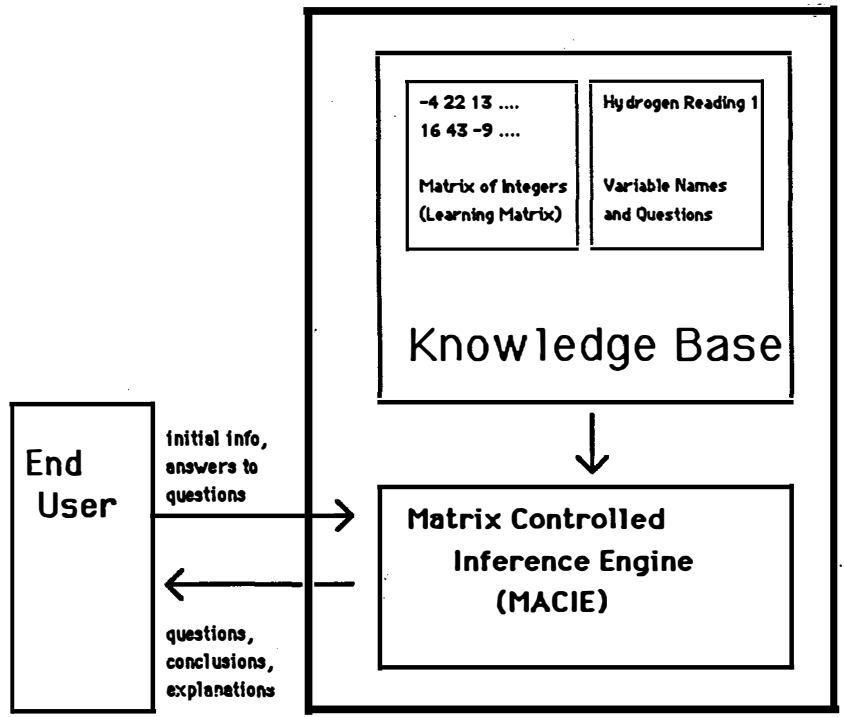

Figure 1: MACIE Style Expert System

Here we will be concerned with a fourth approach that is based upon connectionist network models. For general background consult [Rumelhart & McClelland 1986] and see James A. Anderson's work in [Hinton & Anderson 1981] for some early work in connectionist expert systems.

The particular connectionist expert system approach we will be examining is the MACIE process [Gallant 1985a,Gallant 1985b] which features a connectionist network as a knowledge base. This network is represented internally by a matrix of integers called a *learning matrix* (see figure 1). A MAtrix Controlled Inference Engine (hence the name MACIE) uses this learning matrix for deducing information (forward chaining), seeking out key unknown information (backward chaining), and justifying its conclusions. It is amusing to note that conclusions are justified by giving out IF–THEN rules, even though there are no such rules in the knowledge base. The learning matrix (connectionist network) can be generated from training examples using a number of techniques [Gallant 1986b, Gallant 1986c, Gallant & Smith 1987, Rumelhart & McClelland 1986].

Space limitations preclude a full description of network generation or connectionist expert system operation here.

In this paper we show how the MACIE system can be combined with a deep model and a corresponding noise model to generate an expert system for a problem involving noisy data. The basic idea is to use both models to dynamically generate one training example for every iteration of the program that constructs a connectionist network knowledge base. We will use a small but difficult fault diagnosis problem to illustrate the approach.

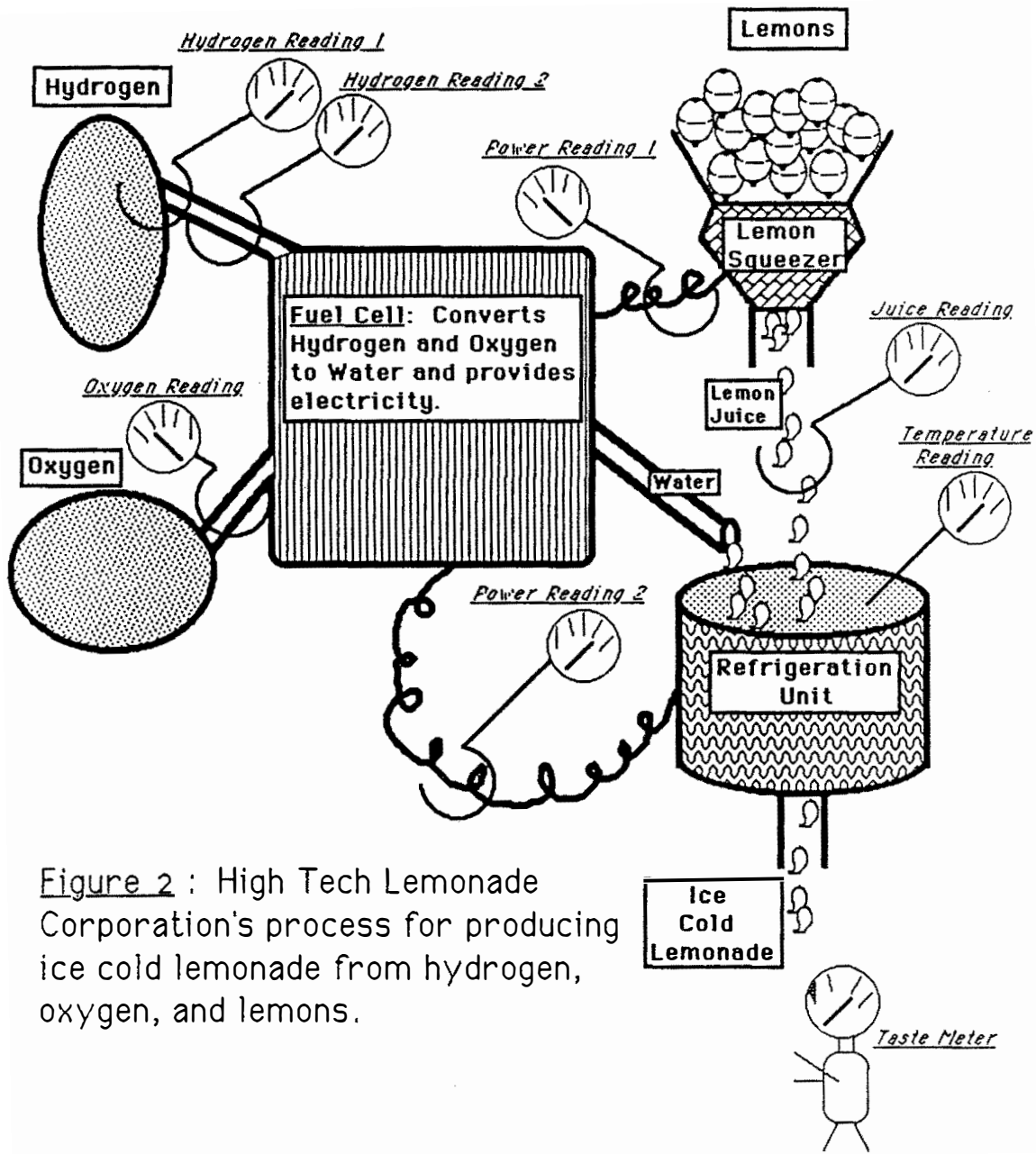

Figure 2 : High Tech Lemonade Corporation's process for producing ice cold lemonade from hydrogen, oxygen, and lemons.

## 2  The High Tech Lemonade Corporation's Problem

High Tech Lemonade Corp. produces ice cold lemonade from three ingredients: hydrogen, oxygen, and lemons (see figure 2). A fuel cell combines the gasses to produce electricity and water. The water goes into the lemonade while the electricity powers a lemon squeezer and a refrigeration unit to cool the resulting drink.

There are a number of failure modes for this somewhat dangerous process (figure 3) and a variety of measurements that are used to diagnose problems (figure 4). Measurements are noisy and redundant and not always available (since the people who monitor the gauges sometimes take several hours off to sit and drink ice cold lemonade).

Suppose we estimate that for 80% of the times that the system is examined it will be operating correctly and when it doesn't work there will be a single failure mode. Suppose also that the lemon squeezer (G7) and refrigeration unit (G8) are twice as likely to fail as other failure modes. Finally let us assume that it is 20 times as important to detect a hydrogen gas problem (G1), and 2 times as important to detect other failures (G2-G8), as it is to detect correct operation of the system (G9).

Our objective is to build an expert system that determines whether any part is not functioning normally based upon whatever data is at hand. For example, our operator might know that Hydrogen Reading 2 indicates a problem and the Temperature Reading is too low but that the lemonade tastes OK. What can he or she conclude? What additional information would be useful? What if that information were not available?

It would be tedious and somewhat difficult to tackle this particular case analytically. To produce a decision tree analysis for all sets of initial information would compare with cleaning the Augean stables.

Building an expert system using a conventional rule-based approach is possible, but it would be difficult to build a non-brittle system. For example, there would be a tendency to generate a rule for hydrogen failure of the form:

> If one (or both??) hydrogen readings indicate a problem then conclude there is a hydrogen problem (with certainty factor ???).

Such a rule would be tricky to construct and debug. It would also be brittle since it would fail to take into account the other readings that are affected by hydrogen problems. These other readings might indicate that there was no hydrogen problem after all, just noisy hydrogen instrument readings.

It is also not clear how a hand-crafted rule should take into account frequency information and relative importance. For example if the importance of diagnosing a hydrogen problem is raised high enough, the system should *always* conclude that such a problem exists. (Even if all readings are normal, there is a small chance that there is a hydrogen problem and all readings are noisy.)

High Tech Lemonade Corporation truly has a problem on its hands.

## 3  The Deep Model and the Noise Model

We tackle this problem by first constructing a deep model for the system that captures all information except for noise. For each failure mode, we list the corresponding meter measurements (assuming no noise) as in figure 5.

For example, a failure in power to the refrigeration unit (G6) would show up on Power Reading 2 (V6) and cause the temperature to rise (V7). Other readings would be normal.

The next step is to construct a set of training examples from figure 5 where entries are duplicated to reflect their relative frequencies. Thus we use two copies of failure mode examples for

G1: Hydrogen Gas Problem
G2: Oxygen Gas Problem
G3: Fuel Cell Not Working
G4: Water Line From Fuel Cell Clogged
G5: Short in Power Line 1 to Lemon Squeezer
G6: Short in Power Line 2 to Refrigeration Unit
G7: Lemon Squeezer Malfunction
G8: Refrigeration Unit Malfunction
G9: All Systems Operating Correctly

Figure 3: Failure Modes for Lemonade Process. Each mode
corresponds to a goal variable G1-G9.

| Variable | Noise |
|---|---|
| V1: Hydrogen Reading 1 | 15% |
| V2: Hydrogen Reading 2 | 25% |
| V3: Oxygen Reading | 20% |
| V4: Power Reading 1 (to Lemon Squeezer) | 15% |
| V5: Lemon Juice Reading | 10% |
| V6: Power Reading 2 (to Refrigeration Unit) | 20% |
| V7: Temperature Reading | 10% |
| V8: Taste Reading (ignoring temperature) | 5% |

Figure 4: Measurements Used for Diagnosing Problems (Variables V1-V8).
A value of True indicates a problem.

| Failure Mode | Frequency | Importance | Final Ratio | V1 | V2 | V3 | V4 | V5 | V6 | V7 | V8 |
|---|---|---|---|---|---|---|---|---|---|---|---|
| G1: | 1 | 20 | 20 | 1 | 1 |   | 1 | 1 | 1 | 1 | 1 |
| G2: | 1 | 2 | 2 |   |   | 1 | 1 | 1 | 1 | 1 | 1 |
| G3: | 1 | 2 | 2 |   |   |   | 1 | 1 | 1 | 1 | 1 |
| G4: | 1 | 2 | 2 |   |   |   |   |   |   |   | 1 |
| G5: | 1 | 2 | 2 |   |   |   | 1 | 1 |   |   | 1 |
| G6: | 1 | 2 | 2 |   |   |   |   |   | 1 | 1 |   |
| G7: | 2 | 2 | 4 |   |   |   |   | 1 |   |   | 1 |
| G8: | 2 | 2 | 4 |   |   |   |   |   |   | 1 |   |
| G9: | 40 | 1 | 40 |   |   |   |   |   |   |   |   |
| Total Examples: |   |   | 78 |   |   |   |   |   |   |   |   |

Figure 5: Frequency data, importance data, and effects of failure modes
on instrument readings (in the absence of noise).
For variables V1-V9 blanks represent values of −1.
The Final Ratio column is the product of frequency and importance.

216

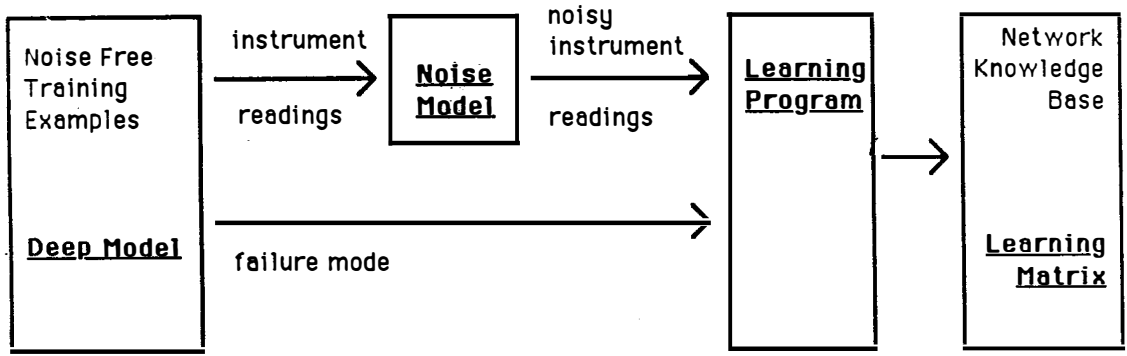

Figure 6: Dynamic Generation of Noisy Training Examples

the squeezer (G7) and refrigeration unit (G8) and 40 copies of the example of correct functioning (G9). Thus a randomly selected example would correspond to a failure mode 10/50 = 20% of the time and would correspond to a correctly functioning system 80% of the time, as desired.

We now make one last change to the set of training examples. We make additional duplicates to represent increased priority of any training examples. For our case, we use a total of 20 copies of the training example for hydrogen supply failure (G1) since we judged this to be 20 times as important as detecting no failure (G9). Similarly we double the copies for other failure modes. The final ratios of training examples are given in figure 5 as the product of frequency and importance.

The reason for duplicating examples is that the learning program that we use to generate the knowledge base seeks to minimize the probability that a randomly selected training example will be misclassified. By duplicating training examples, we effectively adjust the frequency of selection of each particular training example so that the learning program will solve the problem at hand. (Of course the examples do not have to be physically duplicated; they only have to be chosen in accordance with the final ratio column in figure 5.)

The set of training examples we have collected represents the deep model for the problem in the absence of noise. This collection specifies all of the information concerning failure modes, frequencies, and importance.

The noise model has already been given in figure 5. It is the probability that any particular measurement will be incorrect, independently of other measurements. Together these two models allow us to generate noisy training examples that are representative of the problem.

## 4  Generating the Expert System

We modified the learning program to dynamically add noise to training examples as they are randomly selected (figure 6). This was necessary since too many examples would be needed for a static representative set of noisy training examples. Each iteration now proceeds as follows:

1. Pick one of the 78 noise-free training examples at random. It will have exactly one failure mode variable (G1-G9) that is True.

2. Change the value of each *input* variable (V1-V8) from True to False or vice-versa, with probability given in figure 4. Goal variables corresponding to the correct responses (G1-G9) are not changed.

3. Present the modified example to the learning program.

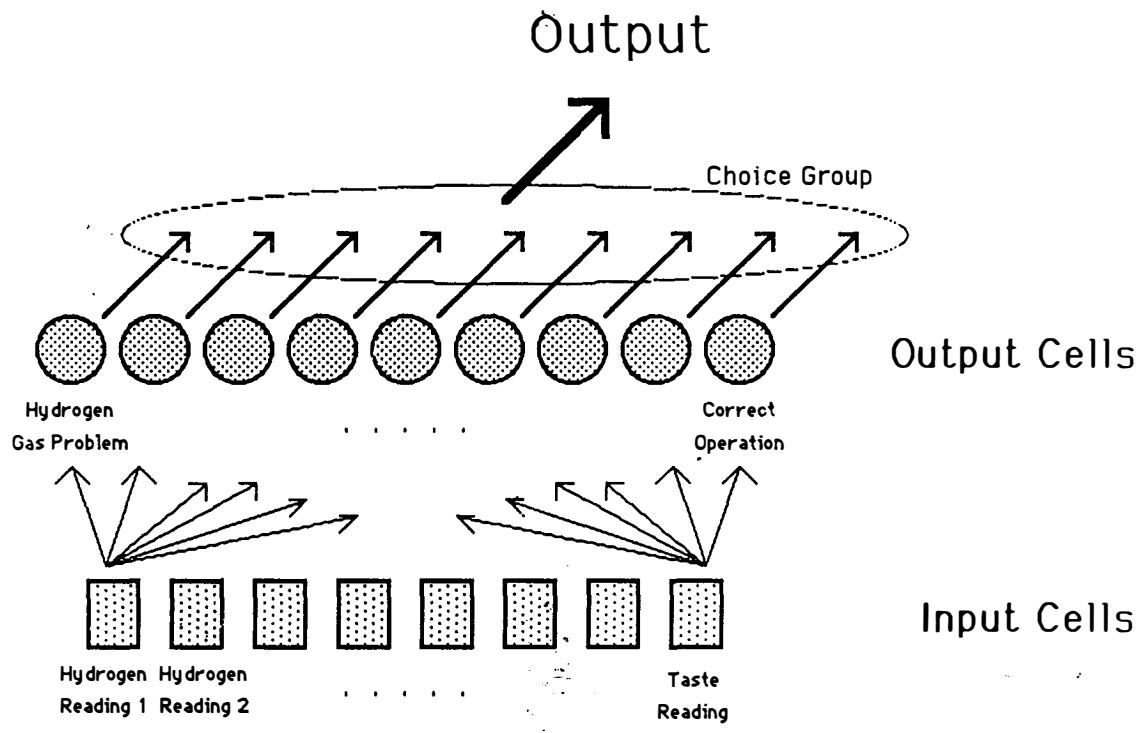

Figure 7: Knowledge Base for Lemonade Fault Detection

The learning program used for this example was the pocket algorithm [Gallant 1986b], a modification of perceptron learning, and it produced a network as illustrated in figure 7. (The connection weights are given in the Appendix in matrix format.) There were 8 input nodes and 9 trainable output cells. The output cells form a choice (or winner-take-all) group where the only cell to fire is the one with the highest weighted combination of its inputs. (This required a small modification of the basic algorithm in [Gallant 1986b].)[1] We used 10,000 iterations to generate the network knowledge base, taking about one minute of connect time. The use of additional intermediate cells as in [Gallant 1986c,Gallant & Smith 1987] was not needed for this problem.

Notice that the resulting system will diagnose a hydrogen problem whenever both of the hydrogen gauges have normal readings and other readings follow the pattern for a hydrogen problem.

We computed a *figure of merit* for the generated system by checking its performance on groups of 1000 noisy training examples randomly selected in accordance with the final ratios of figure 5. The figure of merit averaged approximately 810 out of a possible 1000 points. We compared this with random selection of an output ($1/9 \times 1000 = 111$ points) or with always choosing the most frequently seen output ($40/78 \times 1000 = 513$ points).

We also generated a system without adding noise. This system produced a figure of merit of about 750, significantly worse performance than the model generated with noise. It is interesting to note that the model generated without noise works on every noise-free training example, while the model generated with noise works on only 90%, again evidence that the addition of noise makes a significant difference in the model that is generated.

---

[1] Perceptron learning was generalized so that if choice cell $G_i$ has highest weighted sum for a particular example $E$ when the correct choice was $G_j$, then $G_i$ has $E$ subtracted from its weights and $G_j$ has $E$ added to its weights. Length of a run of correct responses is maintained for the choice group as a whole, and the entire choice group's coefficients are changed during pocket replacement.

Notice that this was a rather noisy problem; a completely noise-free example is generated only about one time in four. For this reason we judged the 810 figure of merit to be a good result. In fact we do not think it would be possible to hand-construct a comparable expert system by conventional means. Moreover it is easy to make changes to the resulting system since regenerating the knowledge base is relatively fast.

## 5  MACIE and Choice Groups of Variables

Any network created by the learning program is immediately available for use as the knowledge base for a connectionist expert system. See [Gallant 1985a] or [Gallant & Balachandra 1986] for an example of a typical session with such a system.

Inferencing with groups of choice variables can be illustrated with the following small example consisting of 3 goal variables (G1–G3) and 3 input variables (V1–V3):

| Choice Variable | Learning Matrix | | | | Current Weighted Sum |
|---|---|---|---|---|---|
| | Constant | V1 | V2 | V3 | |
| G1 | -1 | 2 | 2 | 5 | 1 |
| G2 | -2 | 2 | -1 | -5 | -3 |
| G3 | 1 | 3 | 3 | -4 | 4 |
| | (+1) | ?? | +1 | ?? | ← Current Values |

Currently V2 is known to be True (+1), but V1 and V3 are unknown. G3 dominates G2 since knowing V1 and V3 can change *relative* sums by at most $|2 - 3| + |(-5) - (-4)| = 2$. Therefore we can conclude that G2 is false. At this point the system will ask the user for V3 and conclude G1 if V3 is True and G3 if V3 is false.

## 6  Discussion and Conclusion

The MACIE process, like most other approaches to generating expert systems from training examples, is best suited for classification or diagnostic problems. More procedural or search-oriented tasks such as configuring computers would be poor candidates for such an automated approach. With this qualification, however, MACIE seems generally promising as a tool for constructing expert systems for noisy and redundant domains.

While the sample problem contained only boolean input variables, it should be emphasized that non-boolean variables can be represented by groups of boolean variables [Gallant 1985a].

The techniques are fast enough to generate connectionist expert system knowledge bases for at least medium sized problems and, once a knowledge base is generated, the execution time for the resulting expert system is very fast. Thus the technology appears ready for testing with real applications.

We are currently extending these methods to handle time sequential readings in a more elaborate fault detection model.

# References

[Cheeseman 1983]
Cheeseman, P.C. A Method of Computing Generalized Bayesian Probability Values for Expert Systems, Proc. Eighth International Conference on Artificial Intelligence, Karlsruhe, W. Germany, Aug 8-12, 1983, pp. 198-202

[Cheeseman 1984]
Cheeseman, P.C. Learning of Expert System Data, Proc. IEEE Workshop on Principles of Knowledge Based Systems, Denver, Dec 3-4, 1984, pp. 115-122

[Fisher 1936]
Fisher, R. A. The use of multiple measurements in taxonomic problems. Ann. Eugenics, 7, Part II, 179-188. Also in *Contributions to Mathematical Statistics* (1950) John Wiley, New York.

[Gallant 1985a]
Gallant, S. I. Automatic Generation of Expert Systems From Examples. Proceedings of Second International Conference on Artificial Intelligence Applications, sponsored by IEEE Computer Society, Miami Beach, Florida, Dec. 11-13, 1985.

[Gallant 1985b]
Gallant, S. I. Matrix Controlled Expert System Producible from Examples. Patent Pending 707,458.

[Gallant 1986a]
Gallant, S. I. Brittleness and Machine Learning. International Meeting on Advances in Learning, Les Arcs, France, July 1986; also chapter in forthcoming book edited by Y. Kodratoff and R. Michalski

[Gallant 1986b]
Gallant, S. I. Optimal Linear Discriminants. Proc. Eighth International Conference on Pattern Recognition, Paris, France, Oct. 28-31, 1986.

[Gallant 1986c]
Gallant, S. I. Three Constructive Algorithms for Network Learning. Proc. Eighth Annual Conference of the Cognitive Science Society, Amherst, Ma., Aug. 15-17, 1986.

[Gallant & Balachandra 1986]
Gallant, S. I., & Balachandra, R. Using Automated Techniques to Generate an Expert System for R&D Project Monitoring. International Conference on Economics and Artificial Intelligence, Aix-en-Provence, France, Sept. 2-4, 1986.

[Gallant & Smith 1987]
Gallant, S. I., and Smith, D. Random Cells: An Idea Whose Time Has Come and Gone... And Come Again? IEEE International Conference on Neural Networks, San Diego, Ca., June 1987.

[Hinton & Anderson 1981]
Hinton, G. E. & Anderson, J. A., Eds. *Parallel Models of Associative Memory.* Lawrence Erlbaum Assoc., Hillsdale, N. J. (1981).

[Michalski 1983]
Michalski, R. S., Carbonell, J. G., & Mitchell, T. M. *Machine Learning.* (1983) Tioga Pub. Co., Palo Alto, Ca.

[Michalski 1986]
Michalski, R. S., Carbonell, J. G. & Mitchell, T. M. *Machine Learning, Volume II.* (1986) Morgan Kaufmann Pub., Inc., Los Altos, Ca.

[Minsky & Papert 1969]
Minsky, M. & Papert, S. *Perceptrons: An Introduction to Computational Geometry.* (1969) MIT Press, Cambridge, Ma.

[Pearl 1985]
Pearl, J. How to Do with Probabilities What People Say You Can't. Proceedings of Second

International Conference on Artificial Intelligence Applications, sponsored by IEEE Computer Society, Miami Beach, Florida, Dec. 11-13, 1985.

[Pearl 1986]
Pearl, J. Fusion, Propagation, and Structuring in Belief Networks. Artificial Intelligence 29 (1986), 241-288.

[Rosenblatt 1961]
Rosenblatt, F. *Principles of neurodynamics: Perceptrons and the theory of brain mechanisms.* Spartan Press, Washington, DC.

[Rumelhart & McClelland 1986]
D. E. Rumelhart & J. L. McClelland (Eds.) *Parallel Distributed Processing: Explorations in the Microstructures of Cognition.* MIT Press.

[Stanfill & Waltz 1986]
Stanfill, C. & Waltz, D. Toward Memory Based Reasoning. CACM 29, number 12, December 1986.

[Zadeh 1965]
Zadeh, L. A. Fuzzy Sets. Information and Control 8: 338-353

# 7  Appendix: Learning Matrix

```
G/ Problem with Hydrogen supply
    9    9    5   -3    3    5    3    3    5
G/ Problem with Oxygen supply
   -4   -2   -2    8    2    2    0    2    2
G/ Fuel cell malfunction:  not working
   -2    0    0    4    4    4    4    2    0
G/ Fuel cell malfunction:  water clogged
   -3   -1    1   -1   -5   -7   -1   -3    1
G/ Short in power line to lemon squeezer
   -3   -3   -1   -1    3    1   -3    1    5
G/ Short in power line to refrigeration unit
    0    0    0   -2   -4   -4    4    2   -6
G/ Lemon squeezer malfunction
   -3    1   -1   -3   -5    5    1   -5    5
G/ Refrigeration unit malfunction
   -1   -1    1    1    1   -5   -5    5   -5
G/ System functions correctly
    7   -3   -3   -3    1   -1   -3   -7   -7
  bias  V1   V2   V3   V4   V5   V6   V7   V8
```

```
NOTE:  The first column is a bias constant to be added to the corresponding
       cell's sum.
```